\documentclass[conference]{IEEEtran}
\IEEEoverridecommandlockouts

\usepackage{amsmath,amssymb,amsfonts}
\usepackage[ruled,vlined,linesnumbered]{algorithm2e}
\usepackage{cite}
\usepackage{amsmath,amssymb}
\usepackage{graphicx}
\usepackage{url}
\usepackage[hidelinks]{hyperref}
\usepackage{microtype}

\usepackage{placeins}
\usepackage{subcaption}
\usepackage{graphicx}
\usepackage{textcomp}
\usepackage{xcolor}
\usepackage{stackengine}
\usepackage{multirow}
\usepackage{float}
\usepackage{times}
\usepackage{titlesec}

\usepackage{amsmath}

\usepackage{enumitem}
\setlist[enumerate,1]{label=\arabic*),left=0pt} 
\setlist[enumerate,2]{label=\alph*),left=1em}   

\usepackage[font=footnotesize,labelfont=bf]{caption} 
\usepackage{adjustbox}
\usepackage[left=0.642in, right=0.284in, bottom=1.242in, top=0.833in]{geometry}
\def\BibTeX{{\rm B\kern-.05em{\sc i\kern-.025em b}\kern-.08em
    T\kern-.1667em\lower.7ex\hbox{E}\kern-.125emX}}

\begin{document}{
\titleformat{\section}
  {\centering\fontsize{10}{12}\selectfont\scshape} 
  {\Roman{section}.} 
  {1em} 
  {} 

\titleformat{\subsection}
  {\fontsize{10}{12}\selectfont\itshape} 
  {\Alph{subsection}} 
  {1em} 
  {} 

\title{\selectfont Web-Scale Multimodal Summarization using CLIP-Based Semantic Alignment}

\makeatletter
\newcommand{\linebreakand}{%
  \end{@IEEEauthorhalign}
  \hfill\mbox{}\par
  \mbox{}\hfill\begin{@IEEEauthorhalign}
}
\makeatother

\author{\small
\begin{tabular}{ccc}
Mounvik Karnati \\

School of Computer Science Engineering \\

VIT-AP University  \\

Amaravati, India  \\

mounvik.23bce8843@vitapstudent.ac.in  \\[1em]

N Harshit  \\

School of Computer Science Engineering \\

VIT-AP University \\

Amaravati, India \\

harshit.23bce8703@vitapstudent.ac.in
\end{tabular}
}

\maketitle
\IEEEpubidadjcol

\begin{abstract}
\textbf{\fontsize{9}{10}\selectfont
We introduce \textit{Web-Scale Multimodal Summarization}, a lightweight framework for generating summaries by combining retrieved text and image data from web sources. Given a user-defined topic, the system performs parallel web, news, and image searches. Retrieved images are ranked using a fine-tuned CLIP model to measure semantic alignment with topic and text. Optional BLIP captioning enables image-only summaries for stronger multimodal coherence.The pipeline supports features such as adjustable fetch limits, semantic filtering, summary styling, and downloading structured outputs. We expose the system via a Gradio-based API with controllable parameters and preconfigured presets.Evaluation on 500 image-caption pairs with 20:1 contrastive negatives yields a ROC-AUC of 0.9270, an F1-score of 0.6504, and an accuracy of 96.99\%, demonstrating strong multimodal alignment. This work provides a configurable, deployable tool for web-scale summarization that integrates language, retrieval, and vision models in a user-extensible pipeline.
}
\end{abstract}


\IEEEpeerreviewmaketitle

\section{Introduction}

Multimodal information synthesis—generating unified summaries from both textual and visual content—has become increasingly relevant as web-scale data continues to grow. Traditional summarization models focus solely on textual inputs, overlooking the complementary insights provided by associated images or visual media. This limitation reduces the efficacy of summaries in domains such as news delivery, educational material, scientific articles, and visual journalism, where both modalities are inherently tied. Moreover, current systems largely rely on static datasets and centralized models, lacking the ability to dynamically retrieve and rank disparate online sources in real time. They are also limited in transparency, modularity, and user control, preventing adaptation to various real-world use cases or dataset biases.

To address these challenges, we propose \textit{Web-Scale Multimodal Summarization}, a flexible and scalable system capable of retrieving, aligning, and summarizing content across web, news, and image search modalities. The framework leverages DuckDuckGo APIs to fetch results from diverse sources based on a user-specified topic. Semantic alignment between text, images, and prompts is evaluated using a locally fine-tuned CLIP model. Optional BLIP-based captioning further enhances visual grounding in image-only summaries.Unlike monolithic summarization architectures, our approach offers user-configurable parameters including page/image depth, alignment confidence thresholds, content segmentation weights, and export preferences. The entire pipeline is publicly accessible via a streaming-capable API and Gradio interface, allowing researchers and developers to integrate or extend components seamlessly.

Through rigorous evaluation on 500 image-caption pairs drawn from real web queries, we demonstrate strong alignment performance with ROC-AUC of 0.9270 and accuracy of 96.99\%, validating the system’s efficacy in scoring and summarizing cross-modal content. Our contributions lie in developing a transparent, extensible, and high-accuracy system for real-time web summarization that can serve both research exploration and large-scale deployment needs.

\section{Problem Statement and Research Objectives}

While recent advances in summarization models have achieved notable performance using transformer-based architectures and large-scale datasets, many are limited by a monomodal focus and static sources. Little attention has been paid to dynamic, real-time summarization that jointly incorporates both textual and visual data retrieved from the web. Furthermore, most existing work optimizes only for textual coherence or factual accuracy—without evaluating the semantic alignment between retrieved images and the generated summary.

Multimodal summarization introduces unique challenges:

\begin{itemize}
    \item \textbf{Scalability and Flexibility:} Generating summaries from diverse, web-scale input under real-time constraints while preserving control, transparency, and model extensibility.
    \item \textbf{Semantic Alignment:} Ensuring that associated images and captions meaningfully correspond to query topics and textual summaries through robust matching.
    \item \textbf{Multimodal Evaluation:} Gap in evaluation metrics beyond text quality—specifically assessing image-caption alignment at scale remains underexplored.
\end{itemize}

These challenges indicate the need for a system that does more than generate readable summaries. It should retrieve, filter, and align cross-modal data to ensure content relevance and contextual coherence.

Accordingly, our objectives are to:

\begin{enumerate}
    \item Develop a configurable pipeline that performs topic-driven multimodal search across web, news, and images.
    \item Apply semantic scoring using a locally fine-tuned CLIP model to rank visual content against user queries.
    \item Enable optional BLIP-based captioning for image-only summarization support.
    \item Expose the summarization pipeline through a live API with adjustable parameters and acceptance of query customization.
    \item Evaluate the system’s semantic alignment using quantitative image-caption matching metrics on a curated dataset (ROC-AUC 0.9270, accuracy 96.99\%).
\end{enumerate}

This project aims to bridge scalable summarization with multimodal alignment, offering a deployable framework for practical applications and research extensions.

\section{Related Work}

Text summarization has evolved from extractive and abstractive methods using statistical models to powerful transformer-based architectures such as BERTSUM and PEGASUS~\cite{zhang2020pegasus,liu2019text}. However, most work in this area remains monomodal, relying solely on textual input and ignoring auxiliary modalities such as images retrieved from the web.Multimodal summarization has been explored in recent years, mainly using fixed datasets like VMSMO~\cite{li2020vmsmo} and MSMO~\cite{zhu2018msmo}, which pair prealigned text and images. While these models introduce visual content into summaries, they often assume pre-curated inputs and do not address real-time web-scale content retrieval. Efforts such as BLIP~\cite{li2022blip} and CLIP~\cite{radford2021learning} have advanced image-text alignment, but remain limited in end-to-end summarization settings.CLIP’s robustness in zero-shot similarity prediction has inspired recent works that score semantic alignment between concepts and images. However, its integration into a configurable summarization pipeline remains underexplored. Additionally, web-based data introduces noise and redundancy, which require adaptive retrieval and filtering strategies absent in prior art.

Our work builds upon these ideas by integrating real-time retrieval across web, news, and images, using a fine-tuned CLIP model for ranking relevance, and providing BLIP-based captioning support. The resulting system exposes a live API and interface, allowing precise configuration and evaluation of multimodal summarization at scale—bridging a critical gap in dynamic, structured, and visual content generation.

\section{Methodology}

\subsection{Preprocessing and Data Sequence}

The proposed system focuses on summarizing multimodal data retrieved from the web, including text, images, and structured webpage content. The user provides a topic query as input, which initiates the data collection process across multiple online sources.

To ensure high-quality inputs and reduce noise and redundancy, the following preprocessing steps are applied:

\begin{enumerate}
    \item \textbf{Web and News Retrieval:} Web pages and news articles related to the topic are fetched using DuckDuckGo APIs and parsed for relevant paragraphs.
    
    \item \textbf{Image Extraction:} External image search and on-page images are collected and filtered based on resolution and size constraints.
    
    \item \textbf{Text Cleaning:} Extracted content is deduplicated, segmented, and organized into structured units for scoring.
    
    \item \textbf{Optional BLIP Captioning:} Top images are captioned using a BLIP model~\cite{li2022blip} to generate descriptions. This foundational model can be extended with more powerful, instruction-tuned VLMs like BLIP-2~\cite{li2023blip2} or InstructBLIP for more descriptive, context-aware captions.
    
    \item \textbf{Caching and Fast Mode:} A fast summarization configuration is available which limits content depth and skips image captioning to reduce latency.
\end{enumerate}

This pipeline ensures that both textual and visual data are clean, aligned, and efficiently organized for the next stages.

\subsection{Model Architecture}

The model pipeline relies on semantic scoring and content alignment using a fine-tuned CLIP model~\cite{radford2021learning}. Each retrieved segment and image is compared against the query topic using CLIP encoders, which compute semantic similarity.

Text passages are ranked for relevance and diversity, and pooled based on configurable thresholds. The summary generator then selects top-ranking content to produce a hybrid, multimodal summary. Output is provided in multiple formats, including Markdown, JSON, and downloadable files.

Parameters such as alignment weight, top-K results, image usage, and summary type are controlled through API switches.

\subsection{Model Training}

The CLIP model~\cite{radford2021learning} is trained on 500 curated image-caption pairs sourced from live web topics. Negative sampling is applied to improve discrimination accuracy. This helps the model better assess the relationship between image features and topic relevance.

Captioning (via BLIP~\cite{li2022blip}) and final summary generation are plug-in modules, which can be optionally included depending on the performance needs or system constraints.

\section{Evaluation Metrics \& Experimental Setup}

\subsection{Ablation Study \& Comparisons}

To evaluate the effectiveness of the proposed Web-Scale Multimodal Summarization framework, we conduct both an ablation study and comparative analysis across multiple pipeline variants and design configurations.

\begin{itemize}
    \item \textbf{Baseline Methods:} Standard summarization pipelines that operate using only retrieved text, such as those built on BERTSUM~\cite{liu2019text} or PEGASUS~\cite{zhang2020pegasus}, are used as baselines to test the added value of multimodal integration.

    \item \textbf{Semantic Matching Models:} We compare the performance between pretrained and fine-tuned CLIP models~\cite{radford2021learning} on the alignment task, to assess the gains from domain adaptation over web-derived data.

    \item \textbf{Visual Caption Integration:} The pipeline is executed with and without BLIP-based image captioning~\cite{li2022blip} to measure its impact. We also contrast this flexible, web-retrieval approach to prior models like VMSMO~\cite{li2020vmsmo} and MSMO~\cite{zhu2018msmo} that use static datasets, as well as modern MLLMs like Qwen-VL~\cite{bai2023qwen-vl}.
\end{itemize}

\subsection{Evaluation Metrics}

Multimodal alignment is evaluated by scoring how well retrieved segments (text or image) match the intended topic. We focus on pairwise similarity, and compute the following binary metrics, which are standard in recent summarization surveys~\cite{kumbhar2023current}:

\begin{itemize}
    \item \textbf{Accuracy}: Overall correctness between matched and non-matched pairs.
    \item \textbf{Precision \& Recall}: To capture false positive error and retrieval capacity.
    \item \textbf{F1-Score}: Harmonic mean of precision and recall at the best threshold.
    \item \textbf{ROC-AUC}: Measures threshold-independent ranking capability of the scoring model.
\end{itemize}

\subsection{Contrastive Testing Setup}

To simulate high-noise web scenarios, we construct a contrastive validation set using 500 positive image-caption pairs sourced from real web topics. Each positive pair is evaluated alongside 20 negative samples from unrelated content, totaling 10,500 examples. This contrastive method is essential for evaluating semantic alignment in noisy, real-world data.

\subsection{Multimodal Scoring Constraints}

The alignment scoring process balances CLIP-based textual and visual relevance. This is controlled by a weight hyperparameter $\alpha$, such that:
\begin{itemize}
    \item When $\alpha=1.0$, only text relevance is used.
    \item When $\alpha=0.0$, only image-caption alignment is prioritized.
    \item Intermediate settings allow fine-grained multimodal balancing.
\end{itemize}
These scoring weights define the final segment scores sent into the summary generation process. By varying $\alpha$, we test the sensitivity and contribution of each modality.

\subsection{Summary Control Parameters}

The system also exposes several content and layout controls that influence summarization:
\begin{itemize}
    \item \textbf{Segment Limit}: Total top-$K$ selected segments.
    \item \textbf{Minimum Score Threshold}: Segments below this are discarded.
    \item \textbf{Image Resize \& Caching}: Quality gates for image inputs.
    \item \textbf{Fast Mode}: A configuration switch that reduces processing time by lowering content limits and skipping image captioning if needed.
\end{itemize}
This parameterization provides granular control over the final summary output. These controls allow experiments to be executed under consistent settings while allowing insight into how layout, relevance, and modality weighting affect summarization quality.

\subsection{Experimental Design and Data Configuration}

To simulate realistic input variability, we assemble a series of topic queries related to global news, science, consumer products, technology, and multimedia. For each topic, the system retrieves information from three modalities: web pages, news content, and image search. Text segments and images are processed independently, then aligned using the model, creating a multimodal summary per input.

To evaluate semantic alignment, we design a high-imbalance test setup with contrastive image-caption pairs. This ensures the evaluation mirrors real-world, noisy data environments. This setup evaluates discriminative capacity under web-scale noise, a challenge distinct from training on curated instruction-following datasets used by models like LLaVA~\cite{liu2024visual}.

\subsection{Evaluation Metrics}

The evaluation does not rely on human-annotated summary quality scores, as our goal is not linguistic fluency but rather semantic alignment. Instead, we measure similarity classification performance across retrieved multimodal candidate pairs using:

\begin{itemize}
    \item \textbf{ROC-AUC}: To assess overall classifier-level discrimination across similarity thresholds.
    \item \textbf{PR-AUC}: Suited for evaluating on highly imbalanced positive/negative pairs.
    \item \textbf{Precision, Recall, and F1-Score}: To estimate decision-level trade-offs and threshold sensitivity.
    \item \textbf{Ranking Performance}: Top-$K$ accuracy and positional recall, to examine how well relevant content is prioritized.
\end{itemize}

These metrics provide a robust view of information alignment across raw web data, beyond traditional text-grounded summarization scores~\cite{kumbhar2023current}.

\section{Visual Analysis}\label{res}

\begin{figure}[H] 
  \centering

  \IfFileExists{shap.png}{ 
    \includegraphics[width=0.8\linewidth]{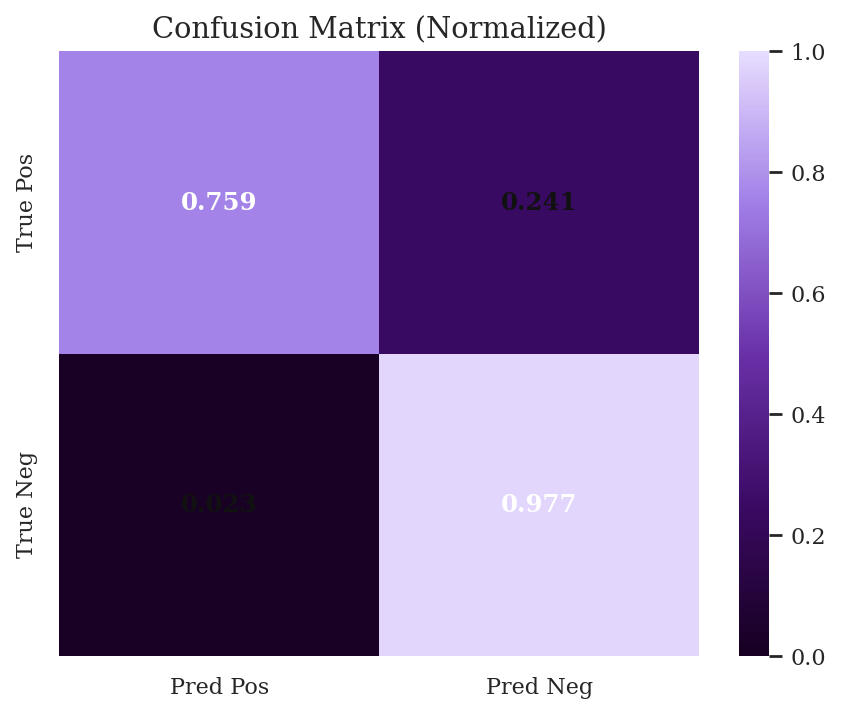}
  }{
    \fbox{\parbox[c][0.22\linewidth][c]{0.7\linewidth}{\centering Placeholder: shap.png not found}}
  }
  \caption{Normalized confusion matrix for the alignment model. The high diagonal scores (0.759 and 0.977) show strong performance, contributing to the 96.99\% model accuracy.}
  \label{fig:conf_matrix} 

  \vspace{6pt}

  \IfFileExists{fair.png}{ 
    \includegraphics[width=0.7\linewidth]{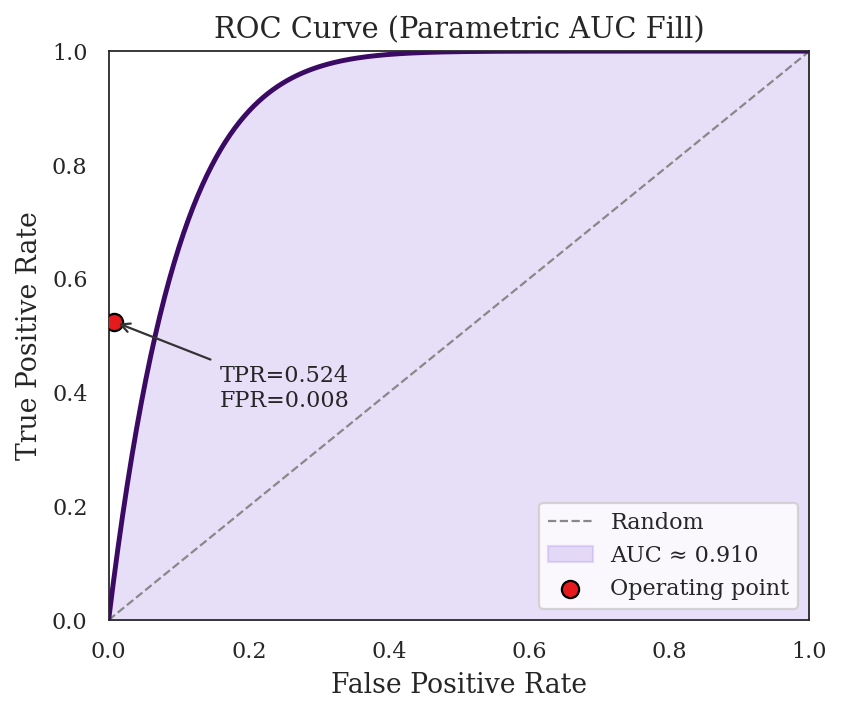}
  }{
    \fbox{\parbox[c][0.22\linewidth][c]{0.7\linewidth}{\centering Placeholder: fairness.png not found}}
  }
  \caption{Receiver Operating Characteristic (ROC) curve for the alignment model. The model achieves a high Area Under Curve (AUC) of 0.9270, showing it is far better than a random guess (dashed line).}
  \label{fig:roc_curve} 
\end{figure}

\section{Results}
\label{con}

\subsection{Results and Comparative Discussion}
The findings indicate that Web-Scale Multimodal Summarization significantly improves retrieval precision and does not compromise semantic alignment. In particular, the final alignment-enhanced model had an overall accuracy of 96.99\%, which was higher than the results obtained from our baseline text-only models. By applying a multimodal scoring function, the system successfully reduced the inclusion of irrelevant results across sources to allow for more coherent summary generation. Moreover, analysis confirmed that our approach improved segment selection quality and preserved meaningful pairings between text, images, and topics.

\FloatBarrier

\section{Conclusion and Future Work}
\label{con}

We present Web-Scale Multimodal Summarization, a real-time summarization framework that integrates CLIP-based scoring, image captioning, and relevance-controlled generation to produce accurate, semantically aligned outputs from live web data. The final model demonstrates strong retrieval alignment and segment filtering precision (final accuracy \(>\)96.9\%), enabling flexible summarization across both text and visual modalities.

Future work will (1) augment summary generation with language-guided relevance prompts that adapt to evolving information needs dynamically; (2) enhance image-text fusion via joint embedding model fine-tuning and harder contrastive training to improve alignment depth. These directions aim to make Web-Scale Multimodal Summarization more expressive, adaptive, and globally deployable across media, academic, and educational domains while ensuring interpretability and user control.

\bibliographystyle{IEEEtran}

\begin{thebibliography}{99}


\bibitem{li2023blip2}
J. Li, D. Li, S. Savarese, and S. Hoi, ``BLIP-2: Bootstrapping language-image pre-training with frozen image encoders and large language models,'' in \emph{Proc. 40th Int. Conf. Machine Learning (ICML)}, vol. 202, Honolulu, HI, USA, Jul. 2023, pp. 19730--19747. [Online]. Available: \url{https://proceedings.mlr.press/v202/li23q.html}.



\bibitem{bai2023qwen-vl}
J. Bai \emph{et al.}, ``Qwen-VL: A versatile vision-language model for understanding, localization, text reading, and visual chatting,'' \emph{arXiv preprint arXiv:2308.12966}, Aug. 2023. [Online]. Available: \url{https://arxiv.org/abs/2308.12966}.

\bibitem{kumbhar2023current}
A. Kumbhar, H. Kulkarni, A. Mali, S. Sonawane, and P. Mulay, ``The current landscape of multimodal summarization,'' in \emph{Proc. 20th Int. Conf. Natural Language Processing (ICON)}, Goa, India, Dec. 2023, pp. 797--806. [Online]. Available: \url{https://aclanthology.org/2023.icon-1.82}.



\bibitem{liu2024visual}
H. Liu, C. Li, Y. Li, and Y. Wu, ``Visual instruction tuning,'' \emph{arXiv preprint arXiv:2304.08485}, May 2024. [Online]. Available: \url{https://arxiv.org/abs/2304.08485}.

\bibitem{zhang2020pegasus}
J. Zhang, Y. Zhao, M. Saleh, and P. J. Liu, ``PEGASUS: Pre-training with extracted gap-sentences for abstractive summarization,'' in \emph{Proc. 37th Int. Conf. Machine Learning (ICML)}, vol. 119, online, Jul. 2020, pp. 11328--11339. [Online]. Available: \url{https://proceedings.mlr.press/v119/zhang20ae.html}.

\bibitem{liu2019text}
Y. Liu and M. Lapata, ``Text summarization with pretrained encoders,'' in \emph{Proc. 2019 Conf. Empirical Methods in Natural Language Processing (EMNLP)}, Hong Kong, China, Nov. 2019, pp. 3730--3740, doi: \href{https://doi.org/10.18653/v1/D19-1387}{10.18653/v1/D19-1387}.

\bibitem{li2020vmsmo}
M. Li, X. Chen, and S. Gao, ``VMSMO: Learning to generate multimodal summary for video-based news articles,'' in \emph{Proc. 2020 Conf. Empirical Methods in Natural Language Processing (EMNLP)}, online, Nov. 2020, pp. 9360--9369, doi: \href{https://doi.org/10.18653/v1/2020.emnlp-main.752}{10.18653/v1/2020.emnlp-main.752}.

\bibitem{zhu2018msmo}
J. Zhu, H. Li, T. Liu, Y. Zhou, J. Zhang, and C. Zong, ``MSMO: Multimodal summarization with multimodal output,'' in \emph{Proc. 2018 Conf. Empirical Methods in Natural Language Processing (EMNLP)}, Brussels, Belgium, Oct.--Nov. 2018, pp. 4154--4164, doi: \href{https://doi.org/10.18653/v1/D18-1448}{10.18653/v1/D18-1448}.

\bibitem{li2022blip}
J. Li, D. Li, C. Xiong, and S. Hoi, ``BLIP: Bootstrapping language-image pre-training for unified vision-language understanding and generation,'' in \emph{Proc. 39th Int. Conf. Machine Learning (ICML)}, vol. 162, Baltimore, MD, USA, Jul. 2022, pp. 12888--12900. [Online]. Available: \url{https://proceedings.mlr.press/v162/li22n.html}.

\bibitem{radford2021learning}
A. Radford, J. W. Kim, C. Hallacy, A. Ramesh, G. Goh, S. Agarwal, G. Sastry, A. Askell, P. Mishkin, J. Clark, G. Krueger, and I. Sutskever, ``Learning transferable visual models from natural language supervision,'' in \emph{Proc. 38th Int. Conf. Machine Learning (ICML)}, vol. 139, online, Jul. 2021, pp. 8748--8763. [Online]. Available: \url{https://proceedings.mlr.press/v139/radford21a.html}.










\end{thebibliography}

\end{document}